\documentclass[runningheads]{llncs}
\usepackage{graphicx}

\usepackage{tikz}
\usepackage{comment} 
\usepackage{amsmath,amssymb} 
\usepackage{color}

\usepackage{setspace}
\usepackage{booktabs}
\usepackage{float}
\usepackage{subfigure}
\usepackage{wrapfig}
\usepackage{multicol}
\usepackage{multirow}
\usepackage{booktabs}
\usepackage{setspace}
\usepackage{rotating}
\usepackage{makecell}
\usepackage{ctable}
\usepackage{marvosym}

\usepackage{hyperref}
\hypersetup{
    colorlinks=true,
    linkcolor=blue,
}

\begin{document}
\pagestyle{headings}
\mainmatter
\def\ECCVSubNumber{100}  
\newcommand\blfootnote[1]{%
	\begingroup
	\renewcommand\thefootnote{}\footnote{#1}%
	\addtocounter{footnote}{-1}%
	\endgroup
}
\title{Cascade Graph Neural Networks for RGB-D Salient Object Detection} 

\titlerunning{{\scshape{Cas-Gnn}} for RGB-D Salient Object Detection}
%
\author{Ao Luo\inst{1}$^{\star}$ \and
Xin Li\inst{2}$^{\star}$ \and
Fan Yang\inst{2}\and
Zhicheng Jiao\inst{3}\and \\
Hong Cheng\inst{1}\textsuperscript{\Letter}\and
Siwei Lyu\inst{4}}

\authorrunning{A. Luo, et al.}
%
\institute{
Center for Robotics, School of Automation Engineering, UESTC, Chengdu, China\and
Group 42 (G42), Abu Dhabi, UAE\and
University of Pennsylvania, Philadelphia, USA\and
University at Albany, State University of New York, USA\\
\email{\{aoluo,xinli,fanyang\}\_uestc@hotmail.com};  \email{hcheng@uestc.edu.cn} 
}

\maketitle

\begin{abstract}

		\blfootnote{$\star$ Equal contribution}\blfootnote{{\Letter} Corresponding author}In this paper, we study the problem of salient object detection (SOD) for RGB-D images using both color and depth information. A major technical challenge in performing salient object detection from RGB-D images is how to fully leverage the two complementary data sources. Current works either simply distill prior knowledge from the corresponding depth map for handling the RGB-image or blindly fuse color and geometric information to generate the coarse depth-aware representations, hindering the performance of RGB-D saliency detectors. In this work, we introduce {\em Cascade Graph Neural Networks} ({\scshape{Cas-Gnn}}), a unified framework which is capable of comprehensively distilling and reasoning the mutual benefits between these two data sources through a set of cascade graphs, to learn powerful representations for RGB-D salient object detection. {\scshape{Cas-Gnn}} processes the two data sources individually and employs a novel {\em Cascade Graph Reasoning} (CGR) module to learn powerful dense feature embeddings, from which the saliency map can be easily inferred. Contrast to the previous approaches, the explicitly modeling and reasoning of high-level relations between complementary data sources allows us to better overcome challenges such as occlusions and ambiguities. Extensive experiments demonstrate that {\scshape{Cas-Gnn}} achieves significantly better performance than all existing RGB-D SOD approaches on several widely-used benchmarks. Code is available at \url{https://github.com/LA30/Cas-Gnn}.

\end{abstract}

	\section{Introduction}

Salient object detection is the crux to dozens of high-level AI tasks such as object detection or classification~\cite{ren2013region,zhang2017bridging,xie2019attentive}, weakly-supervised semantic segmentation~\cite{jin2017webly,wang2018weakly}, semantic correspondences~\cite{Yang_2017_CVPR} and others~\cite{li2017visual,xu2015show,xie2019srsc}. An ideal solution should identify salient objects of varying shape and appearance, show robustness towards heavy occlusion, various illumination and background. With the development of hardware (sensors and GPU), prediction accuracy of data-driven methods that use deep networks~\cite{Zhao_2019_ICCV,Liu_2019_ICCV,Xu_2019_ICCV,Su_2019_ICCV,Zeng_2019_ICCV,Wu_2019_ICCV,Wu_2019_CVPR,Zhang_2019_CVPR,Fan_2019_CVPR,Chen_2018_ECCV,Li_2018_ECCV} have been improved significantly, compared to traditional methods based on hand-crafted features~\cite{liu2010learning,cheng2014global,zhang2013saliency,zhang2015minimum}. However, these approaches only take the appearance features from RGB data into consideration, making them unreliable when handling the challenging cases, such as poorly-lighted environments and low-contrast scenes, due to the lack of depth information.

The depth map captured by RGB-D camera preserves important geometry information of the given scene, allowing 2D algorithms to be extend into 3D space. Depth awareness has been proven to be crucial for many applications of scene understanding, \emph{e.g.,} scene parsing~\cite{wang2018depth,Jiao_2019_CVPR}, 6D object pose estimation~\cite{wang2019densefusion,he2019pvn3d} and object detection~\cite{gupta2014learning,qi2018frustum}, leading to a significant performance enhancement. Recently, there have been a few attempts to take into account the 3D geometric information for salient object detection in the given scene, \emph{e.g.,} by distilling prior knowledge from the depth~\cite{ren2015exploiting} or incorporating depth information into a SOD framework~\cite{Zhao_2019_CVPR,Piao_2019_ICCV,fan2019rethinking}. These RGB-D models have achieved better performances than RGB-only models in salient object detection when dealing with challenging cases. However, as we demonstrate empirically, existing RGB-D salient object detection models fall short under heavy occlusions and depth image noise. One primary reason is that these models, which only focus on delivering or gathering information, ignore {modeling and reasoning over high-level  relations} between two data sources. Therefore, it is hard for them to fully exploit the complementary nature of 2D color and 3D depth information for overcoming the ambiguities in complex scenes. These observations inspire us to think about: \textit {How to explicitly reason on high-level relations over 2D appearance (color) and 3D geometry (depth) information for better inferring salient regions}? 

Graph neural network ({{GNN}}) has been shown to be an optimal way of relation modeling and reasoning~\cite{Shen_2018_ECCV,Chen_2019_CVPR,Zhao_2019_CVPR1,xu2018powerful,Wang_2019_ICCV}. Generally, a GNN model propagates messages over a graph, such that the node's representation is not only obtained from its own information but also conditioned on its relations to the neighboring nodes. It has revolutionized deep representation learning and benefitted many computer vision tasks, such as 3D pose estimation~\cite{Cai_2019_ICCV}, action recognition~\cite{Zhao_2019_ICCV_graph}, zero-shot learning~\cite{xie2020} and language grounding~\cite{Bajaj_2019_ICCV}, by incorporating graph computation into deep learning frameworks. However, how to design a suitable GNN model for RGB-D based SOD is challenging and, to the best of our knowledge, is still unexplored.	

In this paper, we present the first attempt to build a GNN-based model, namely {\em Cascade Graph Neural Networks} ({\scshape{Cas-Gnn}}), to explicitly reason about the 2D appearance and 3D geometry information for RGB-D salient object detection. Our proposed deep model including multiple graphs, where each graph is used to handle a specific level of \emph{cross-modality} reasoning.  In each graph, two basic types of nodes are contained, \emph{i.e.,} {\tt geometry nodes} storing depth features and  {\tt appearance nodes} storing RGB-related features, and they are linked to each other by edges. Through message passing, the useful mutual information and high-level relations between two data sources can be gradually distilled for learning the powerful dense feature embeddings, from which the saliency map can be inferred. To further enhance the capability for reasoning over multiple levels of features, we make our {\scshape{Cas-Gnn}} to have these multi-level graphs sequentially chained by coarsening the preceding graph into two domain-specific {\tt guidance nodes} for the following cascade graph. Consequently, each graph in our {\scshape{Cas-Gnn}} (except for the first cascade graph) has three types of nodes in total, and they distill useful information from each other to build powerful feature representations for RGB-D based salient object detection. 

Our {\scshape{Cas-Gnn}} is easy to implement and end-to-end learnable.  As opposed to prior works which simply fuse features of the two data sources, {\scshape{Cas-Gnn}} is capable of explicitly reasoning about the 2D appearance and 3D geometry information over chained graphs, which is essential to handle heavy occlusions and ambiguities. Extensive experiments show that our {\scshape{Cas-Gnn}} performs remarkably well on $7$ widely-used datasets, outperforming state-of-the-art approaches by a large margin. In summary, our major contributions are described below:

\begin{enumerate}
	\item[1)] We are the first to use the graph-based techniques to design network architectures for RGB-D salient object detection. This allows us to fully exploit the mutual benefits between the 2D appearance and 3D geometry information for better inferring salient object(s).

	
	\item[2)] We propose a graph-based, end-to-end trainable model, called {\em Cascade Graph Neural Networks} ({\scshape{Cas-Gnn}}), for RGB-D based SOD, and carefully design {\em Graph-based Reasoning} (GR) module to distill useful knowledge from different modalities for building powerful feature embeddings.


	\item[3)] Different from most GNN-based approaches, our {\scshape{Cas-Gnn}} ensembles a set of cascade graphs to reason about relations of the two data sources hierarchically. This cascade reasoning capability ensures the graph-based model to exploit rich, complementary information from multi-level features, which is useful in capturing object details and overcoming ambiguities.

	\item[4)] We conduct extensive experiments on $7$ widely-used datasets and show that our {\scshape{Cas-Gnn}} sets new records, outperforming state-of-the-art approaches.

\end{enumerate}	

	\section{Related Work}
		This work is related to RGB-D based salient object detection, graph neural network and network cascade. Here, we briefly review these three lines of works.

		\noindent{\bfseries \small RGB-D Salient Object Detection.} 
		Unlike approaches for RGB-only salient object detection methods~\cite{Yan_2019_ICCV,Feng_2019_CVPR,Liu_2019_CVPR,Wu_2019_CVPR,Zhang_2019_CVPR,Fan_2019_CVPR,luo2020webly,Fan_2018_ECCV,Fan_2019_CVPR1,Li_2018_ECCV,liu2010learning,cheng2014global,zhang2013saliency,zhang2015minimum} which only focus on 2D appearance feature learning, RGB-D based SOD approaches~\cite{Zhao_2019_CVPR,Piao_2019_ICCV,fan2019rethinking} take two different data sources, \emph{i.e.,} 2D appearance (color) and 3D geometry (depth) information, into consideration. Classical approaches extract hand-crafted features from the input RGB-D data and perform cross-modality feature fusion by various strategies, such as random forest regressor~\cite{song2017depth} and minimum barrier distance~\cite{wang2017rgb}.  However, with handcrafting of features, classic RGB-D based approaches are limited in the expression ability. Recent works such as CPFP~\cite{Zhao_2019_CVPR} integrates deep feature learning and cross-modality fusion within a unified, end-to-end framework. Piao~\emph{et al.}~\cite{Piao_2019_ICCV} futher enhance the cross-modality feature fusion through a recurrent attention mechanism.  Fan~\emph{et al.}~\cite{fan2019rethinking} introduce a depth-depurator to filter out noises in the depth map for better fusing cross-modality features. These approaches, despite the success, are not able to fully reason the high-order relations of cross-modality data, making them unreliable when handling challenges such as occlusions and ambiguities. In comparison, our {\scshape{Cas-Gnn}} considers a better way to distill the mutual benefit of the two data sources by modeling and reasoning their relations over a set of cascade graphs, and we show that such cross-modality reasoning boosts the performance significantly.

		\noindent{\bfseries \small Graph Neural Networks.} 
		In recent years, a wide variety of graph neural network (GNN) based models~\cite{duvenaud2015convolutional,defferrard2016convolutional,scarselli2008graph,Li_2019_ICCV} have been proposed for different applications~\cite{Shen_2018_ECCV,Chen_2019_CVPR,Zhao_2019_CVPR1,Bi_2019_ICCV,luo2020hybrid}. Generally, a GNN can be viewed as a message passing algorithm, where representations for nodes are iteratively computed conditioned on their neighboring nodes through a differentiable aggregation function. Some typical applications in computer vision include semantic segmentation~\cite{Qi_2017_ICCV}, action recognition~\cite{Wang_2018_ECCV}, point cloud classification and segmentation~\cite{wang2019dynamic}, to name a few. In the context of RGB-D based salient object detection -- the task that we study in this paper -- a key challenge in applying GNNs comes from how the graph model learns high-level relations and low-level details simultaneously. To solve this problem, unlike existing graph models, we ensemble a set of sequentially chained graphs to form a unified, cascade graph reasoning model. Therefore, our {\scshape{Cas-Gnn}} is able to reason about relations across multiple feature levels to capture important hierarchical information for RGB-D based SOD, which is significantly different from all existing GNN based models.

		\noindent{\bfseries \small Network Cascade.}  
		Network cascade is an effective scheme for a variety of high-level vision applications. Popular examples of cascaded models include DeCaFA for face alignment~\cite{Dapogny_2019_ICCV}, BDCN for edge detection~\cite{He_2019_CVPR}, Bidirectional FCN for object skeleton extraction~\cite{yang2018multi}, and Cascade R-CNN for object detection~\cite{Cai_2018_CVPR}, to name a few. The core idea of network cascade is to ensemble a set of models to handle challenging tasks in a \emph{coarse-to-fine} or \emph{easy-to-hard} manner. For salient object detection in RGB-only images, only a few attempts employ the network cascade scheme. Li~\emph{et al.}~\cite{li2017multi} use a cascade network for gradually integrating saliency prior knowledge from coarse to fine. Wu~\emph{et al.}~\cite{Wu_2019_CVPR} design a cascaded partial decoder to enhance the learned features for salient object detection. Different from these approaches, our {\scshape{Cas-Gnn}} propagates the knowledge learned from a more global view to assist fine-grained reasoning by chaining multiple graphs, which aids a structured understanding of complex scenes.

				\begin{figure*}[pt]
			\begin{center}
				\includegraphics[width=0.95\linewidth]{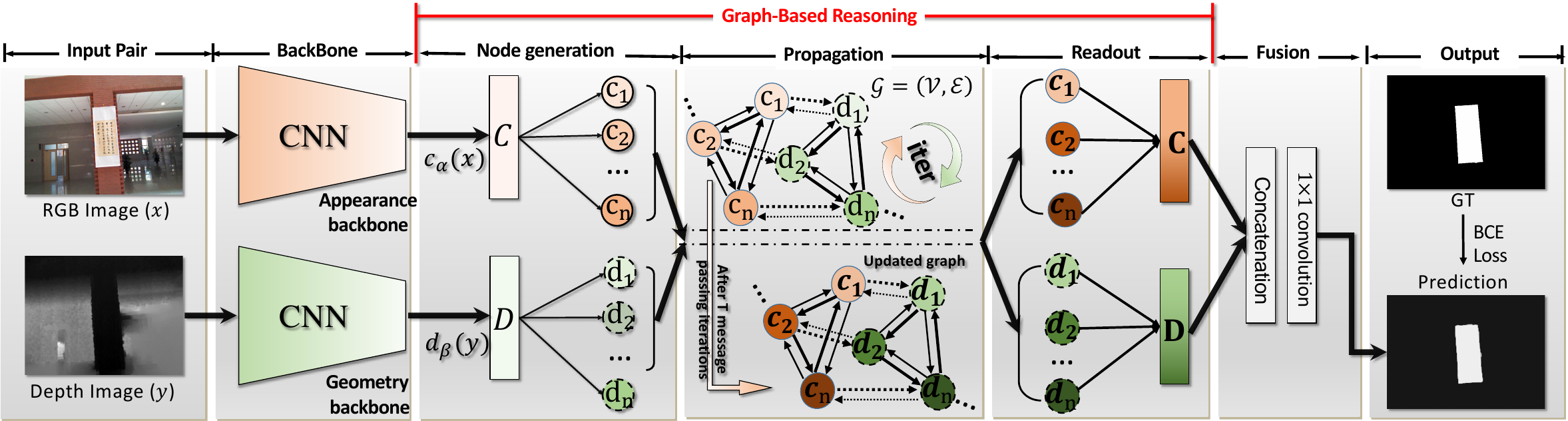}
			\end{center}
			\caption{Overall of our simple cross-modality reasoning model. Our model is built upon two VGG-16 based backbones, and uses a novel graph-based reasoning (GR) module to reason about the high-level relations between the generated 2D appearance and 3D geometry nodes for building more powerful representations. The updated node representations from two modalities are finally fused to infer the salient object regions.}
			
			\label{fig:1}
			
		\end{figure*}
	
\section{Method}
The key idea of {\scshape{Cas-Gnn}} is that it enables the fully harvesting of the 2D appearance and 3D geometric information by using a differentiable, cascade module to hierarchically reason about relations between the two data sources. In this section, we elaborate on how to design a graph reasoning module and how to further enhance the capability of graph-based reasoning using the network cascade technique.  

	\subsection{Problem Formulation}

The task of RGB-D based salient object detection is to predict a saliency map $z \in \mathcal{Z}$ given an input image $x \in \mathcal{X}$ and its corresponding depth image $y \in \mathcal{Y}$. The input space $\mathcal{X}$ and $\mathcal{Y}$ correspond to the space of images and depths respectively, and the target space $\mathcal{Z}$ consists of only one class.  A regression problem is characterized by a continuous target space. In our approach, a graph-based model is defined as a function $f_{\Theta} : \{\mathcal{X},  \mathcal{Y}\} \mapsto \mathcal{Z}$, parameterized by ${\Theta}$, which maps an input pair, \emph{i.e.,} $x \in \mathcal{X}$ and $y \in \mathcal{Y}$, to an output $f_{\Theta} (x,y) \in \mathcal{Z}$. The key challenging is to design a suitable model $\Theta$ that can fully exploit useful information from the two data sources (color and depth image) to learn powerful representations so that it can make the mapping more accurately. 

	\subsection{Cross-modality Reasoning with Graph Neural Networks}
		We start out with a simple GNN model, which reasons over the cross-modality relations between 2D appearance (color) and 3D geometric (depth) information across multiple scales, for salient object detection, as shown in Fig.~\ref{fig:1}.

		\noindent{\bfseries \small Overview.} For RGB-D salient object detection, the key challenge is to fully mine useful information from the two complementary data sources, \emph{i.e.,} the color image $x \in \mathcal{X}$ and the depth $y \in \mathcal{Y}$, and learn the mapping function $f_{\Theta}(x,y)$ which can infer the saliency regions $z \in \mathcal{Z}$. Aiming to achieve this goal, we represent the extracted multi-scale color features $C = \{c_1,\cdots, c_n\}$ and depth features $D = \{d_1,\cdots, d_n\}$ with a directed graph $\mathcal G = (\mathcal{V, E})$, where $\mathcal V$ means a finite set of nodes and $\mathcal E$ stands for the edges among them. The nodes in the GNN model are naturally grouped into two types: the {\tt geometry nodes} $\mathcal V_1 = \{c_1,\cdots,c_n\}$ and the {\tt appearance nodes} $\mathcal V_2 = \{d_1,\cdots, d_n\}$, where $\mathcal V = \mathcal V_1 \cup \mathcal V_2$. The edges $\mathcal{E}$ connect {\bfseries \small i)} the nodes from the same modality ($\mathcal V_1$ or $\mathcal V_2$), and {\bfseries \small ii)} the nodes of the same scale from different modalities, \emph{i.e.,} $c_i \leftrightarrow d_i$ where $i \in \{1,\cdots, n\}$. For each node, $c_i$ or $d_i$, we learn its updated representation, namely $\mathbf c_i^{(t)}$ or $\mathbf d_i^{(t)}$, by aggregating the representations of its neighbors. In the end, the updated features are fused to produce the final representations for salient object detection.

		\noindent{\bfseries \small Feature Backbones.}
		Before reasoning the cross-modality relations, we first extract the 2D appearance feature $\mathcal C$ and 3D geometry feature $\mathcal D$ through the appearance backbone network $c_{\alpha}$ and geometry backbone network $d_{\beta}$, respectively. Following most of the previous approaches~\cite{Piao_2019_ICCV,Chen_2018_CVPR,chen2019multi,han2017cnns,zhu2019pdnet}, we take two VGG-16 networks as the backbones, and use the dilated network technique~\cite{yu2015multi} to ensure that the last two groups of VGG-16 have the same resolution. For the input RGB image $x$ and the corresponding depth image $y$, we can map them to semantically powerful 2D appearance representations $\mathcal C = c_{\alpha}(x)\in \mathbb{R}^{h\times w \times C}$ and 3D geometry representations $\mathcal D = d_{\beta}(y) \in \mathbb{R}^{h\times w \times C}$. Rather than directly fusing the extracted features $\mathcal C$ and $\mathcal D$ to form the final representations for RGB-D salient object detection, we introduce a {\em Graph-based Reasoning} (GR) module to reason about the cross-modality, high-order relations between them to build more powerful embeddings, from which the saliency map can be inferred more easily and accurately. 
		
		    \begin{figure*}[pt]
			\begin{center}
				\includegraphics[width=0.95\linewidth]{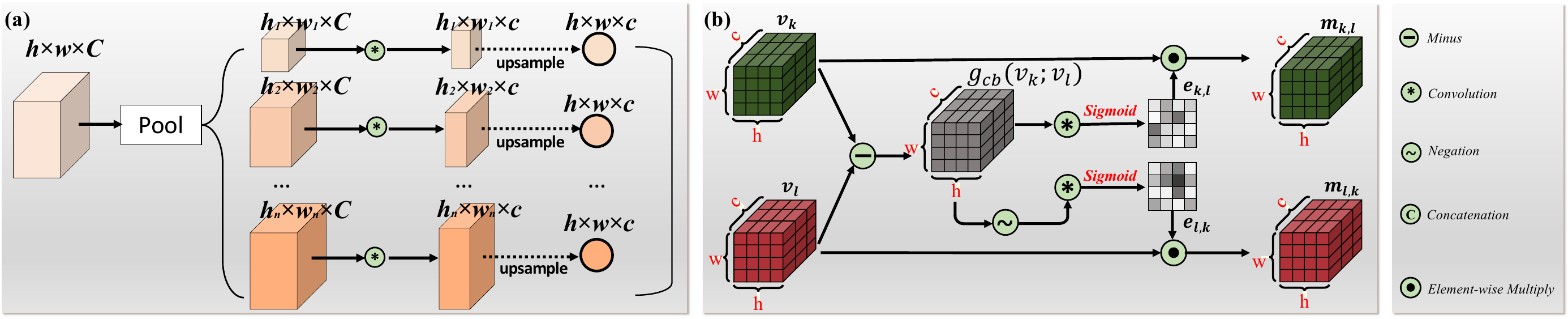}
			\end{center}
			\caption{Detailed illustration of our designs for (a) node embedding and (b) edge embedding. Zoom in for details.}
			\label{fig:k}
		\end{figure*}

		\noindent{\bfseries \small Graph-based Reasoning Module.} 
		The {\em Graph-based Reasoning} (GR) module $g_{\chi}$ takes the underlying 2D appearance features $\mathcal C$ and 3D geometry features $\mathcal D$ as inputs, and outputs powerful embeddings ${\mathbf C}$ and ${\mathbf D}$ after performing cross-modality reasoning: $\{\mathbf C, \mathbf D\} = g_{\chi}(\mathcal C, \mathcal D)$. We formulate $g_{\chi}(\cdot, \cdot)$ in a graph-based, end-to-end differentiable way as follows:
		
\noindent \emph{\textbf{\footnotesize 1) Graph Construction}}: Given the 2D appearance features $\mathcal C$ and 3D geometry features $\mathcal D$, we build a graph $\mathcal G = ({\mathcal V, \mathcal E})$ which has two types of nodes: the {\tt geometry nodes} $\mathcal V_1 = \{c_1,\cdots,c_n\}$ and the {\tt appearance nodes} $\mathcal V_2 = \{d_1,\cdots, d_n\}$, where $\mathcal V = \mathcal V_1 \cup \mathcal V_2$. Each node $c_i$ or $d_i$ is a feature map for a predefined scale $s_i$ and edges link {\bfseries \small i)} the nodes from the same modality but different scales, \emph{i.e.,} $c_i \leftrightarrow c_j$ or $d_i \leftrightarrow d_j$, and {\bfseries \small ii)} the nodes of the same scale from different modalities, \emph{i.e.,} $c_i \leftrightarrow d_i$. Next, we show how to parameterize the nodes $\mathcal V$, edges $\mathcal E$, and message passing functions $\mathcal M$ of the graph $\mathcal G$ with neural networks.

	\noindent \emph{\textbf{\footnotesize 2) Multi-scale Node Embeddings $\mathcal V$}}: Given the 2D appearance features $\mathcal C$ and 3D geometry features $\mathcal D$, as shown in Fig.~\ref{fig:k}(a), we leverage the pyramid pooling module (PPM)~\cite{zhao2017pyramid} followed by a convolution layer and an interpolation layer to extract multi-scale features of the two modalities ($n$ scales) as the initial node representations, resulting in $N = 2\cdot n$ nodes in total. For the appearance node $c_i$ and geometry node $d_i$, their initial node representations $\mathbf c^{(0)}_i \in \mathbb{R}^{h\times w \times c}$ and $\mathbf d^{(0)}_i \in \mathbb{R}^{h\times w \times c}$ can be computed as:
\begin{equation}
\begin{aligned}
\mathbf c^{(0)}_i = \mathcal{R}_{h\times w}(Conv(\mathcal {P}(\mathcal C; s_i))); \quad \mathbf d^{(0)}_i = \mathcal{R}_{h\times w}(Conv(\mathcal {P}(\mathcal D; s_i))),  
\label{eq2}
\end{aligned}
\end{equation}

\noindent where $\mathcal {P}(\cdot~; s_i)$ means the pyramid pooling operation, which pools the given feature maps to the scale of $s_i$, and $\mathcal{R}(\cdot)$ is the interpolation operation which ensures multi-scale feature maps to have the same size $h\times w$. 

\setcounter{footnote}{0}
	\noindent \emph{\textbf{\footnotesize 3) Edge Embeddings $\mathcal E$}}: The nodes are linked by edges for information propagation. As mentioned above, in our constructed graph, edges link  {\bfseries \small i)} the nodes from the same modality but different scales, and {\bfseries \small ii)} the nodes of the same scale from different modalities. For simplification, we use $v_k$ and $v_l$, where $v_k$, $v_l\in\mathcal V$, to represent two nodes linked by the edge\footnote{In our formulation, the edges, message passing function and node-state updating function have no concern with the node types, therefore we simply ignore the node type for more clearly describing the \emph{\textbf{3) edge embeddings}}, \emph{\textbf{4) message passing}} and \emph{\textbf{5) node-state updating}}.}. As shown in Fig.~\ref{fig:k}(b), the edge embedding $\mathbf e_{k,l}$ is used to represent the high-level relation on the two sides of the edge from $v_k$ to $v_l$ through a relation function $f_{rel}(\cdot~;~\cdot)$:
\begin{equation}
\begin{aligned}
\mathbf e_{k,l} = f_{rel}(\mathbf v_k;\mathbf v_l) = Conv(g_{cb}(\mathbf v_k;\mathbf v_l)) \in \mathbb{R}^{h\times w \times c},
\label{eq3}
\end{aligned}
\end{equation}

\noindent where $\mathbf v_k$ and $\mathbf v_l$ are node embeddings for nodes $v_k$ and $v_l$ respectively, $g_{cb}(\cdot~;~\cdot)$ is a function that combines the node embeddings $\mathbf v_k$ and $\mathbf v_l$, and $Conv(\cdot)$ is the convolution operation which learns the relations in an end-to-end manner. For the combination function $g_{cb}(\cdot~;~\cdot)$, we follows~\cite{wang2019dynamic} and model it as: $g_{cb}(\mathbf v_k;\mathbf v_l) = \mathbf v_l - \mathbf v_k$. The resulting edge embedding $\mathbf e_{k,l}$ for node $v_k$ to $v_l$ is also a $c$-dimensional feature map with the size of $h \times w$, in which each feature reflects the pixel-wise relationship between linked nodes. 

\noindent \emph{\textbf{\footnotesize 4) Message Passing $\mathcal M$}}:
In our GNN model, each node aggregates feature messages from all its neighboring nodes. For the message $\mathbf m_{k,l}$ passed from all neighboring nodes $v_k$ to $v_l$, we define the following message passing function $\mathcal M(\cdot~;~\cdot)$:
\begin{equation}
\begin{aligned}
\mathbf m^{(t)}_{k,l} = \sum_{k \in \mathcal N{(l)}} \mathcal M (\mathbf v^{(t-1)}_k,  \mathbf e^{(t-1)}_{k,l}) = \sum_{k \in \mathcal N{(l)}} sigmoid (\mathbf e^{(t-1)}_{k,l})\cdot \mathbf v^{(t-1)}_k \in \mathbb{R}^{h\times w \times c}
\label{eq4}
\end{aligned}
\end{equation}

\noindent where $sigmoid(\cdot)$ is the sigmoid function which maps the edge embedding to link weight. Since our GNN model is designed for a pixel-wise task, the link weight between node is represented by a 2D map.

\noindent \emph{\textbf{\footnotesize 5)Node-state Updating ${\mathcal F}_{update}$}}: 
After the $t$\_th message passing step, each node $v_l$ in our GNN model aggregates information from its neighboring nodes to update its orginal feature representations. Here, we model the node-state updating process with Gated Recurrent Unit~\cite{ballas2015delving}, 
\begin{equation}
\begin{aligned}
\mathbf v_l^{(t)} = \sum_{k \in \mathcal N(l)} {\mathcal F}_{update} (\mathbf v_l^{(t-1)}, \mathbf {m}_{k,l}^{(t-1)}) = \sum_{k \in \mathcal N(l)} \mathcal{U}_{GRU}(\mathbf v_l^{(t-1)}, \mathbf {m}_{k,l}^{(t-1)}),
\label{eq5}
\end{aligned}
\end{equation}

\noindent where $\mathcal{U}_{GRU}(\cdot~;~\cdot)$ stands for the gated recurrent unit.

\noindent \emph{\textbf{\footnotesize 6)Saliency Readout $\mathcal{O}$}}: After $T$ message passing iterations, we upsample all updated node embeddings of each modality to the same size through the interpolation layer $R(\cdot)$, and merge them, \emph{i.e.,} $\mathbf V_1 = \{R(\mathbf c^{(T)}_i) \}^n_{i=1}$ and $\mathbf V_2 = \{R(\mathbf d^{(T)}_i) \}^n_{i=1}$, to form the embeddings:
\begin{equation}
\begin{aligned}
\mathbf C = {{\mathcal F}_{merge}} (\mathbf V_1); \quad \mathbf D = {\mathcal F}_{merge} (\mathbf V_2),
\label{eq6}
\end{aligned}
\end{equation}

\noindent where ${\mathcal F}_{merge}(\cdot)$ denotes the merge function which is implemented with a concatenation layer followed by a $3\times 3$ convolution layer. The learned embeddings of each modality can be further fused to form the final representations for RGB-D salient object detection by the following operation:
\begin{equation}
\begin{aligned}
\mathbf S = \mathcal{R}_{H\times W} (\mathcal{O} (\mathbf C, \mathbf D)),
\label{eq7}
\end{aligned}
\end{equation}

\noindent where $\mathcal{O}(\cdot)$ is the readout function that maps the learned representations to the saliency scores. Here, we implement it with a concatenation layer followed by two $1\times 1$ convolution layers; $\mathcal{R}_{H\times W}(\cdot)$ is used to resize the generated results to the same size of input image $H\times W$ through the interpolation operation. 

	\begin{figure*}[pt]
	\begin{center}
		\includegraphics[width=0.95\linewidth]{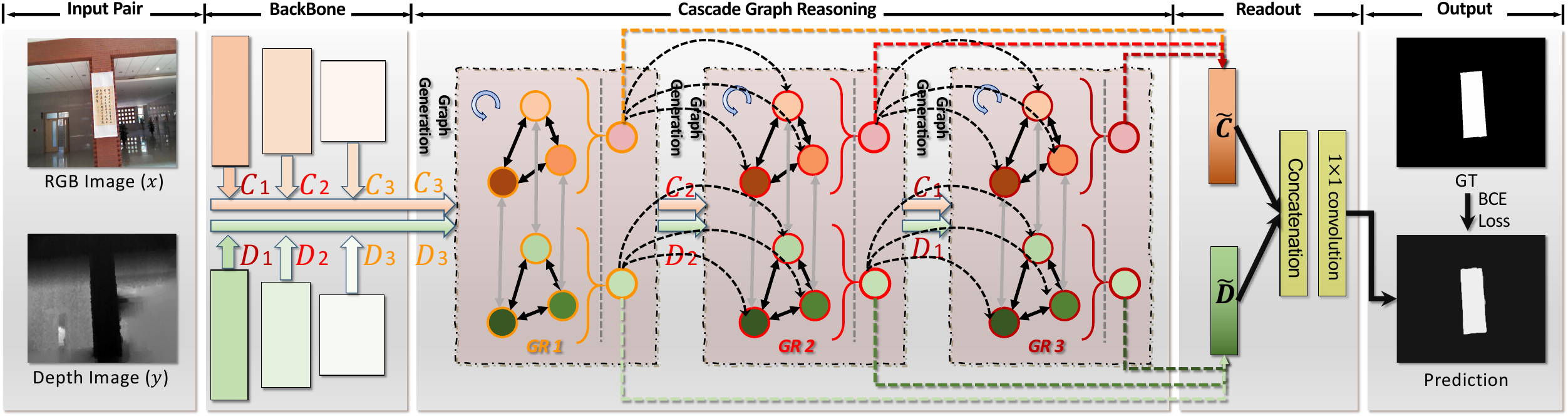}
	\end{center}
	\caption{The overall architecture of our {\scshape{Cas-Gnn}}. Three graph-based reasoning (GR) modules are cascaded in a top-down manner to better distill multi-level information.}
	\label{fig:3}
\end{figure*}

Overall, all components in our GNN model are formulated in a differentiable manner, and thus can be trained end-to-end. Next, we show how to further enhance the capability of GNN model through network cascade techniques. 

	\subsection{Cascade Graph Neural Networks}
In this part, we further enhance our GNN model for RGB-D salient object detection by using the network cascade technique. As observed by many existing works~\cite{hou2017deeply,liu2016dhsnet,wang2016saliency,Fan_2020_CVPR}, the deep-layer and shallow-layer features are complementary to each other: the deep layer features encode high-level semantic knowledge while the shallow-layer features capture rich spatial information. Ideally, a powerful deep saliency model should be able to fully explore these multi-level features. Aiming to achieve this, we extend our GNN model to a hierarchical GNN model which is able to perform the reasoning across multiple levels for better inferring the salient object regions. 

\noindent{\bfseries \small Hierarchical Reasoning via Multi-level Graphs.}
A straightforward scheme is to ensemble a set of graphs across multiple levels $\{\mathcal {G}_w\}_{w=1}^W$ to learn the embeddngs individually, and then fuse the learned representations to build the final representations. Formally,  given the VGG-16 based appearance backbone $c_{\alpha}$ for RGB image $\mathcal X$ and geometry backbone $d_{\beta}$ for depth image $\mathcal Y$, we follow~\cite{hou2017deeply} to map the inputs to $W$ levels of side-output features, \emph{i.e.}, the multi-level appearance features $\Tilde{\mathcal{V}_1} = \{\mathcal C_1,\cdots,\mathcal C_W\}$ and the multi-level geometry features $\Tilde{\mathcal{V}_2} = \{\mathcal D_1,\cdots,\mathcal D_W\}$. For the features of each level $w \in [1, W]$, we build a graph ${\mathcal G}_w$ and use our proposed {\em Graph-based Reasoning} (GR) module $g_{\chi}(\mathcal C_w, \mathcal D_w)$ to map them to the corresponding embeddings $\{\mathbf C_w, \mathbf D_w\}^W_{i=1}$. Then, these multi-level embeddings of each modality, $\Tilde{{V}_1}= \{\mathbf C_1,\cdots,\mathbf C_W\}$ and $\Tilde{{V}_2}= \{\mathbf D_1,\cdots,\mathbf D_W\}$, can be easily interpolated to have the same resolution through the interpolation layer $R(\cdot)$, \emph{i.e.,} $\Tilde{{\mathbf V}_1}= \{R(\mathbf C_1),\cdots,R(\mathbf C_W)\}$ and $\Tilde{{\mathbf V}_2}= \{R(\mathbf D_1),\cdots,R(\mathbf D_W)\}$, and merged by the following function:
\begin{equation}
\begin{aligned}
\Tilde{\mathbf C} = \mathcal{M}_{cl} (\Tilde{\mathbf{V}_1}); \quad \Tilde{\mathbf D} = \mathcal{M}_{cl} (\Tilde{\mathbf{V}_2})
\label{eq8}
\end{aligned}
\end{equation}

\begin{figure}[pt]
	\begin{center}
		\includegraphics[width=0.95\linewidth]{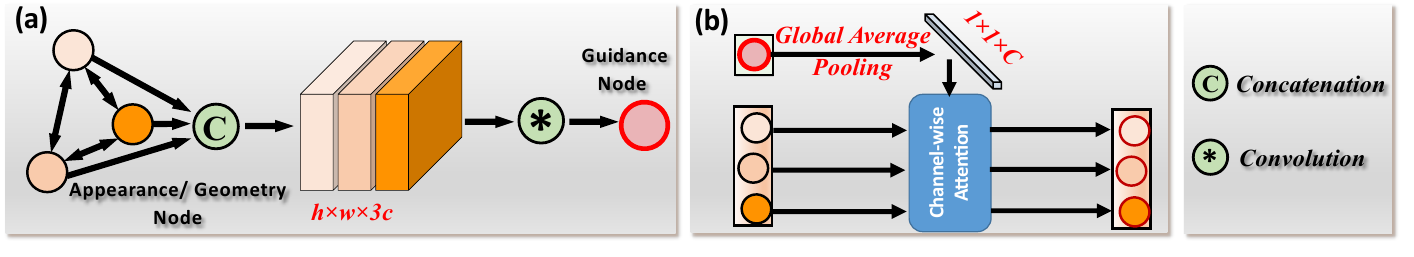}
	\end{center}
	\caption{Detailed illustration of our designs for (a) guidance node generation and (b) attention-based message propagation. Best viewed in color. }
	\label{fig:4}
\end{figure}

\noindent where $\mathcal{M}_{cl}(\cdot)$ is a merge function, which can be either element-wise addition or channel-wise concatenation.  Then, the readout function $\mathcal{O} (\Tilde{\mathbf C}, \Tilde{\mathbf D})$ can be used to generate the final results. 

Generally, this simply hierarchical approach enables the model to perform reasoning across multiple levels. However, as it treats the multi-level reasoning process independently, the mutual benefits are hard to be fully explored. 

\noindent{\bfseries \small Cascade Graph Reasoning.}
To overcome the drawbacks of independent multi-level (graph-based) reasoning, we propose the {\em Cascade Graph Reasoning} (CGR) module by chaining these graphs $\{\mathcal {G}_w\}_{w=1}^W$ for joint reasoning. The resulting model is called {\em Cascade Graph Neural Networks} ({\scshape{Cas-Gnn}}), as shown in Fig.~\ref{fig:3}. Specifically, our {\scshape{Cas-Gnn}} includes multi-level graphs $\{\mathcal{G}_w\}_{w=1}^W$ which are linked in a top-down manner by coarsening the preceding graph into two domain-specific {\tt guidance nodes} for the following cascade graph to perform the joint reasoning.

\noindent \emph{\textbf{\footnotesize 1) Guidance Node}}: Unlike {\tt geometry nodes} and  {\tt appearance nodes}, {\tt guidance nodes} only deliver the guidance information, and will stay fixed during the message passing process. In our formulation, for reasoning the cross-modality relations of the $w$\_th cascade stage, its preceding graph (from the deeper side-output level) is mapped into {\tt guidance node} embeddings by the following functions:
\begin{equation}
\begin{aligned}
\mathbf g^{w}_{c} = \mathcal F(\mathbf V^{(w-1)}_1); \quad \mathbf g^{w}_{d} = \mathcal F(\mathbf V^{(w-1)}_2),  
\label{eq9}
\end{aligned}
\end{equation}

\noindent where	$\mathbf g^{w}_{c}$ and $\mathbf g^{w}_{d}$ are the guidance node embeddings of cascade stage $w$, and ${\mathcal{F}}(\cdot)$ is the graph merging operator, which coarsens the set of learned node embeddings ($\mathbf V^{(w-1)}_1 = \{\mathbf c^{(w-1)(T)}_i \}^n_{i=1}$ or $\mathbf V^{(w-1)}_2 = \{\mathbf d^{(w-1)(T)}_i \}^n_{i=1}$) of the preceding graph $\mathcal G_{(w-1)}$ by firstly concatenating them and then performing the fusion via a $3\times3$ convolution layer (See Fig.~\ref{fig:4}(a)). 


\noindent \emph{\textbf{\footnotesize 2) Cascade Message Propagation}}: Each guidance node,  $\mathbf g^{w}_{c}$ or $\mathbf g^{w}_{d}$, propagates the guidance information to other nodes of the same domain in the graph $\mathcal G_{(w)}$ through the attention mechanism:
\begin{equation}
\begin{aligned}
\Breve {\mathbf v}_c^{w(t)} =  \mathbf v_c^{w(t)} \odot \mathcal A(\mathbf g^{w}_{c}); \quad \Breve {\mathbf v}_d^{w(t)} =  \mathbf v_d^{w(t)} \odot \mathcal A(\mathbf g^{w}_{d})
\label{eq10}
\end{aligned}
\end{equation}

\noindent where $\Breve {\mathbf v}_c^{w(t)}$ and $\Breve {\mathbf v}_d^{w(t)}$ denote the updated appearance node embeddings and geometry node embeddings for the cascade stage $w$ after $t$\_th message passing step respectively; $\odot$ means the channel-wise multiplication. $\mathcal A(\cdot)$ is the attention function, which can be formulated as:
\begin{equation}
\begin{aligned}
A(\mathbf g^{w}_{c}) = sigmoid (\mathcal P(\mathbf g^{w}_{c})); \quad A(\mathbf g^{w}_{d}) = sigmoid (\mathcal P(\mathbf g^{w}_{d})); 
\label{eq11}
\end{aligned}	
\end{equation}

\noindent where $\mathcal P(\cdot)$ is the global average pooling operation, and the $sigmoid$ is used to map the guidance embeddings of each modality to the channel-wise attention vectors (See Fig.~\ref{fig:4}(b)). Therefore, the geometry and appearance node embeddings can incorporate important guidance information from previous graph $\mathcal G_{(w-1)}$ during performing the joint reasoning over $\mathcal G_w$ to create more powerful embeddings. 

\noindent \emph{\textbf{\footnotesize 3) Multi-level Feature Fusion}}: Through the cascade message propagation, the {\em Cascade Graph Reasoning} (CGR) learns the embeddings of multi-level features under the guidance information provided by the guidance nodes. Here, we denote these learned multi-level embeddings as $\{\Breve{\mathbf C_1},\cdots,\Breve{\mathbf C_W}\}$ and $\{\Breve{\mathbf D_1},\cdots,\Breve{\mathbf D_W}\}$. To fuse them, we rewrite Eq.~\ref{eq8} to create the representations:
\begin{equation}
\begin{aligned}
\Breve{\mathbf C} = \mathcal{M}_{cl} (R(\Breve{\mathbf C_1}),\cdots,R(\Breve{\mathbf C_W})); \quad \Breve{\mathbf D} = \mathcal{M}_{cl} (R(\Breve{\mathbf D_1}),\cdots,R(\Breve{\mathbf D_W}));
\label{eq12}
\end{aligned}
\end{equation}

\noindent where $\Breve{\mathbf C}$ and $\Breve{\mathbf D}$ denote the merged representations for the appearance and geometry domain, respectively. Finally, the saliency readout operation (Eq.~\ref{eq7}) is used to produce the final saliency map.

\section{ Experiments}
In this section, we first provide the implementation details of our {\scshape{Cas-Gnn}}. Then, we perform ablation studies to evaluate the effectiveness of each core component of graph-based model. Finally, {\scshape{Cas-Gnn}} is compared with several state-of-the-art RGB-D SOD methods on six widely-used datasets.  

\noindent{\bfseries \small Datasets:} We conduct our experiments on $7$ widely-used datasets: NJUD~\cite{ju2014depth}, STEREO~\cite{niu2012leveraging}, NLPR~\cite{peng2014rgbd},  LFSD~\cite{li2014saliency}, RGBD135~\cite{cheng2014depth}, and SSD~\cite{zhu2017three}. For fair comparison, we follow most SOTAs~\cite{Chen_2018_CVPR,chen2019multi,han2017cnns} to randomly select 1,400 samples from the NJU2K dataset and 650 samples from the NLPR dataset for training, and use all remaining images for evaluation. 

\noindent{\bfseries \small Evaluation Metrics:} We adopt $5$ most-widely used evaluation metrics to comprehensively evaluate the performance of our model, including the mean absolute error ({\tt MAE}), the precision-recall curve ({\tt PR Curve}), F-measure ($F_{\beta}$), S-measure ($S_{\alpha}$)~\cite{fan2017structure} and E-measure($E_{\xi}$)~\cite{fan2018enhanced}. Following previous SOTAs~\cite{Chen_2018_CVPR,chen2019multi,han2017cnns}, we set $\beta$ in $F_{\beta}$ to $0.3$ and $\alpha$ in $S_{\alpha}$ to 0.5 for fair comparison.

\begin{figure}[pt]
	\begin{center}
		\subfigure{\includegraphics[width=0.24\linewidth]{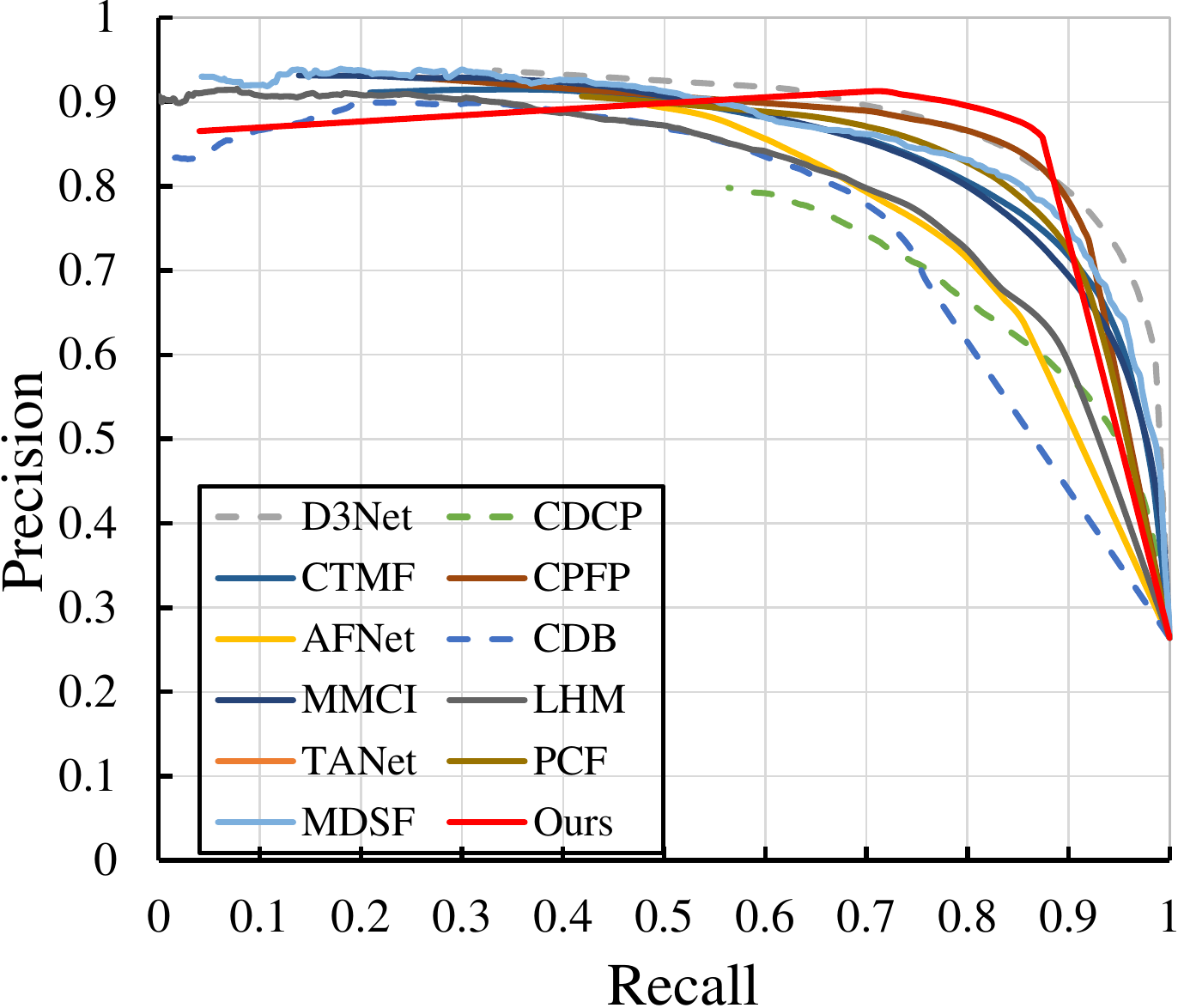}}
		\subfigure{\includegraphics[width=0.24\linewidth]{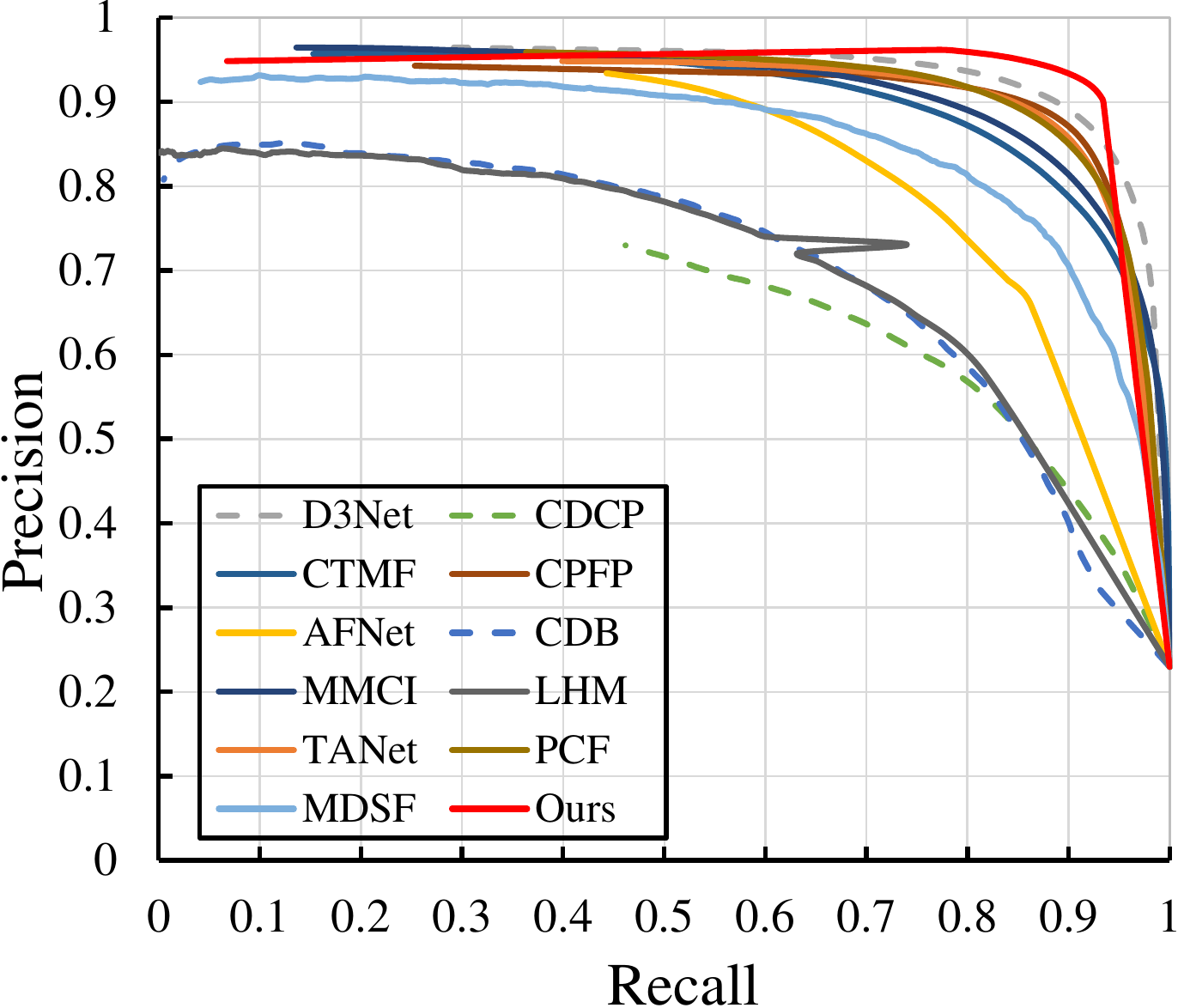}}
		\subfigure{\includegraphics[width=0.24\linewidth]{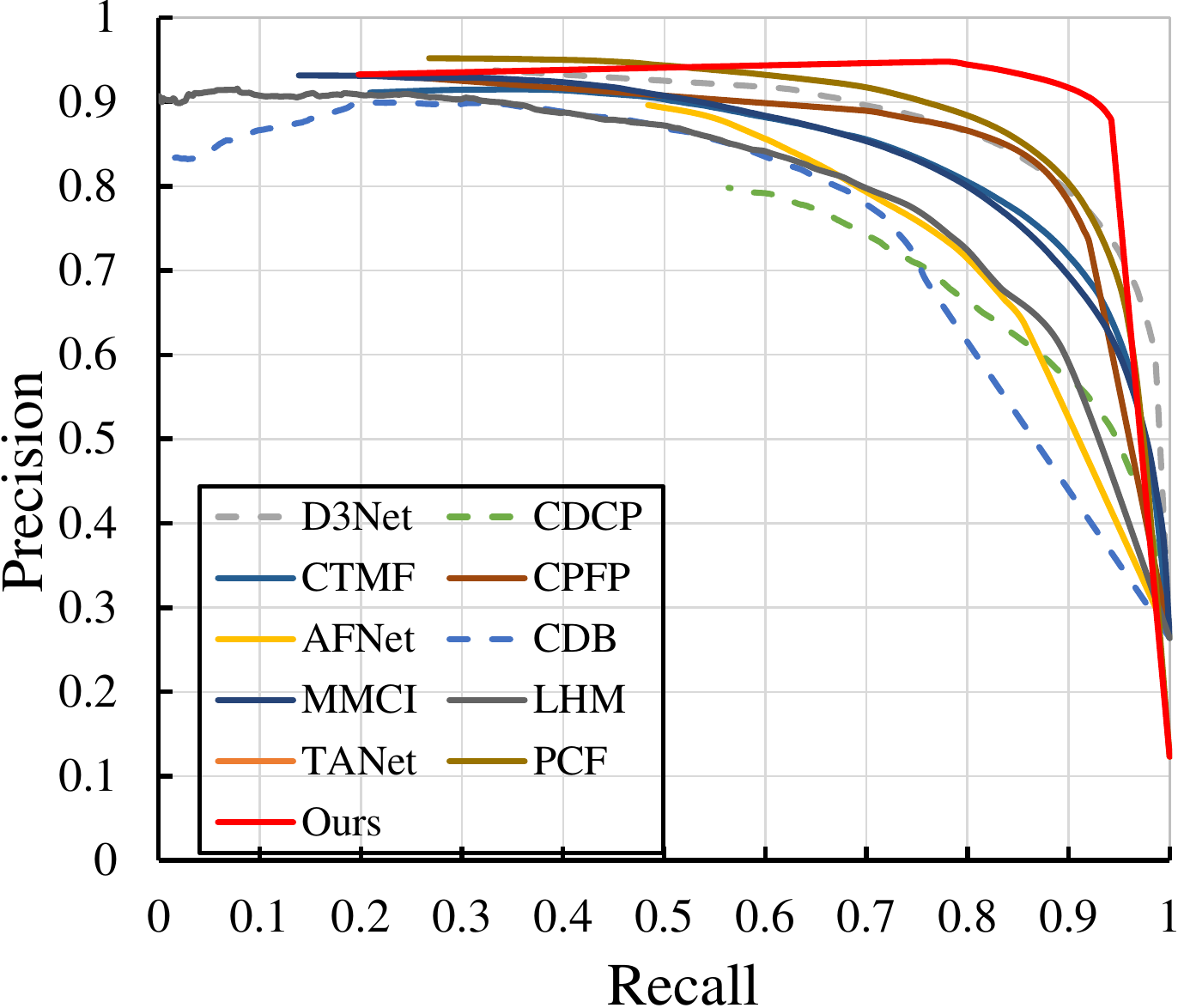}}
		\subfigure{\includegraphics[width=0.24\linewidth]{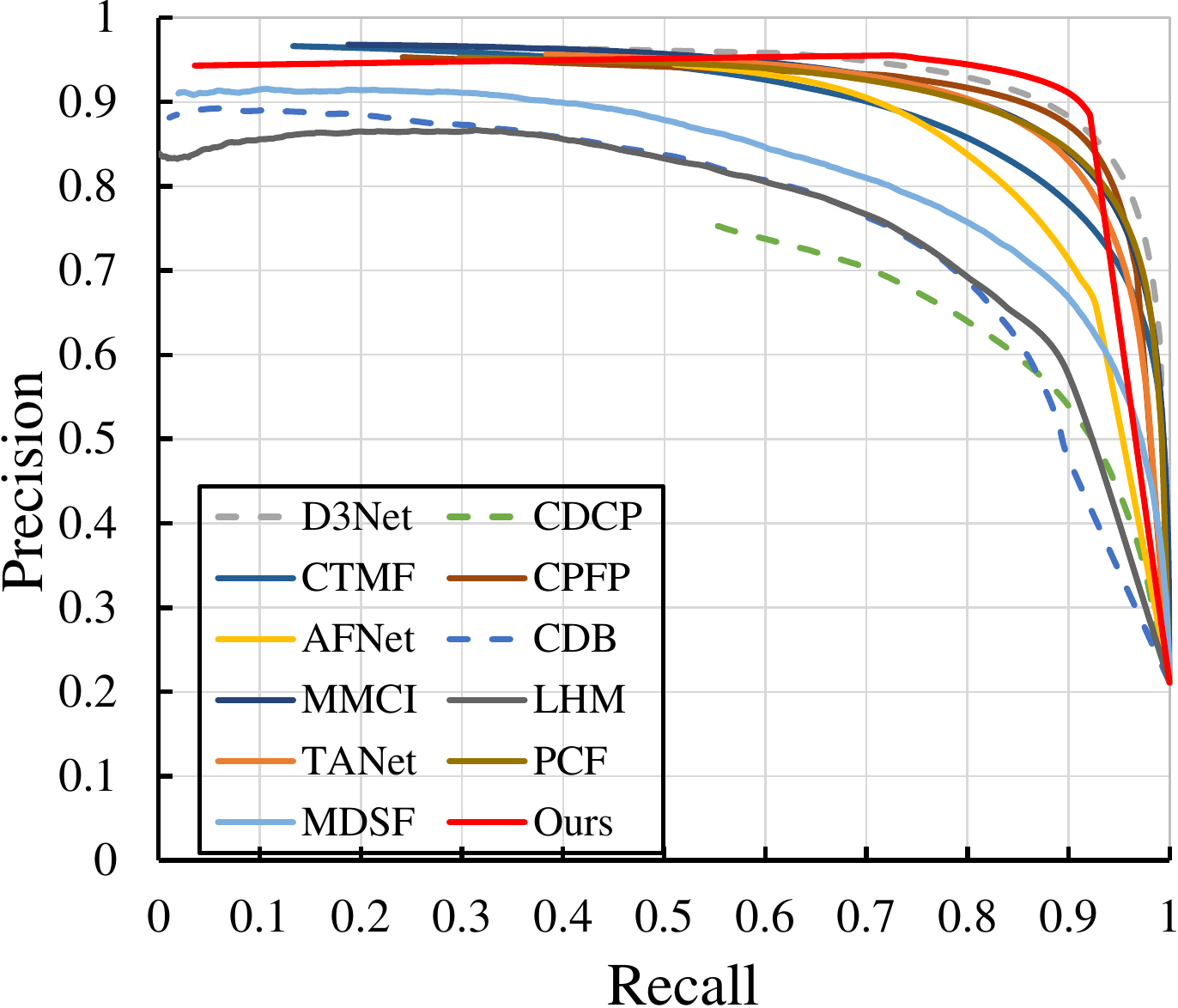}}\\
		\setcounter{subfigure}{0}
		\subfigure[LFSD]{\includegraphics[width=0.235\linewidth]{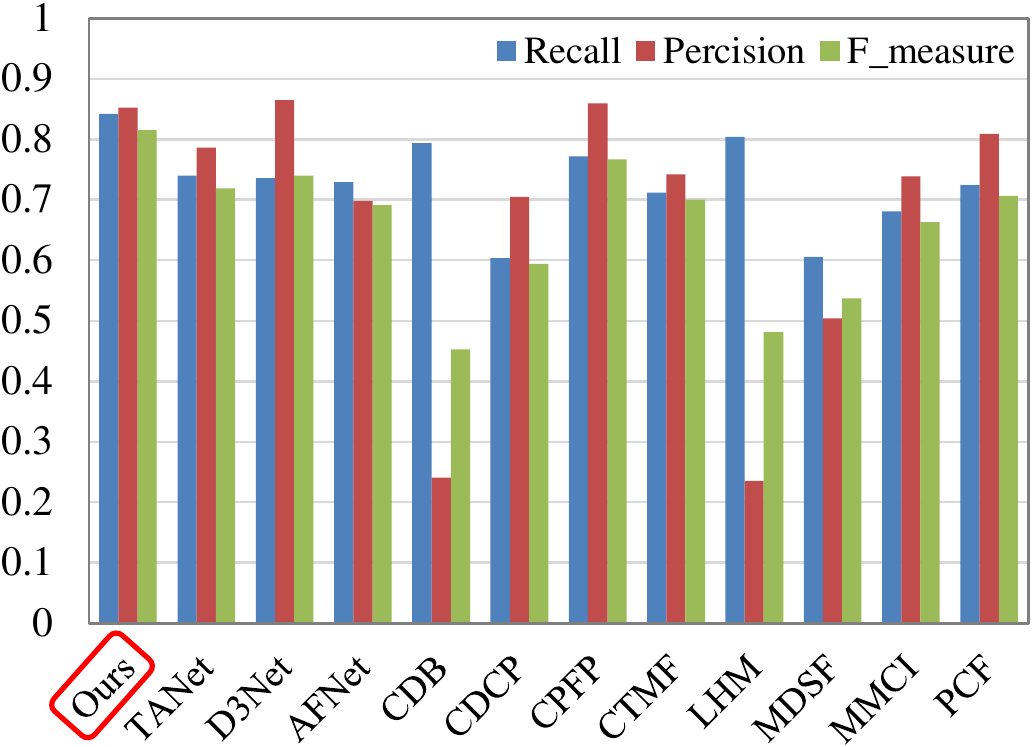}}
		\subfigure[NJUD]{\includegraphics[width=0.235\linewidth]{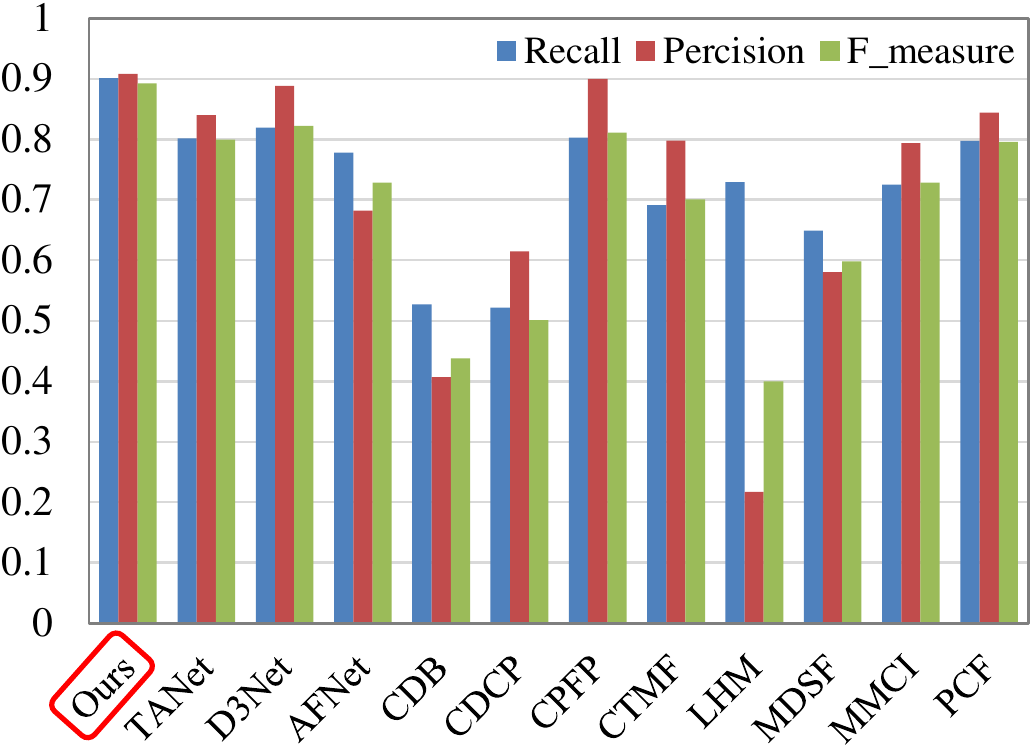}}
		\subfigure[NLPR]{\includegraphics[width=0.235\linewidth]{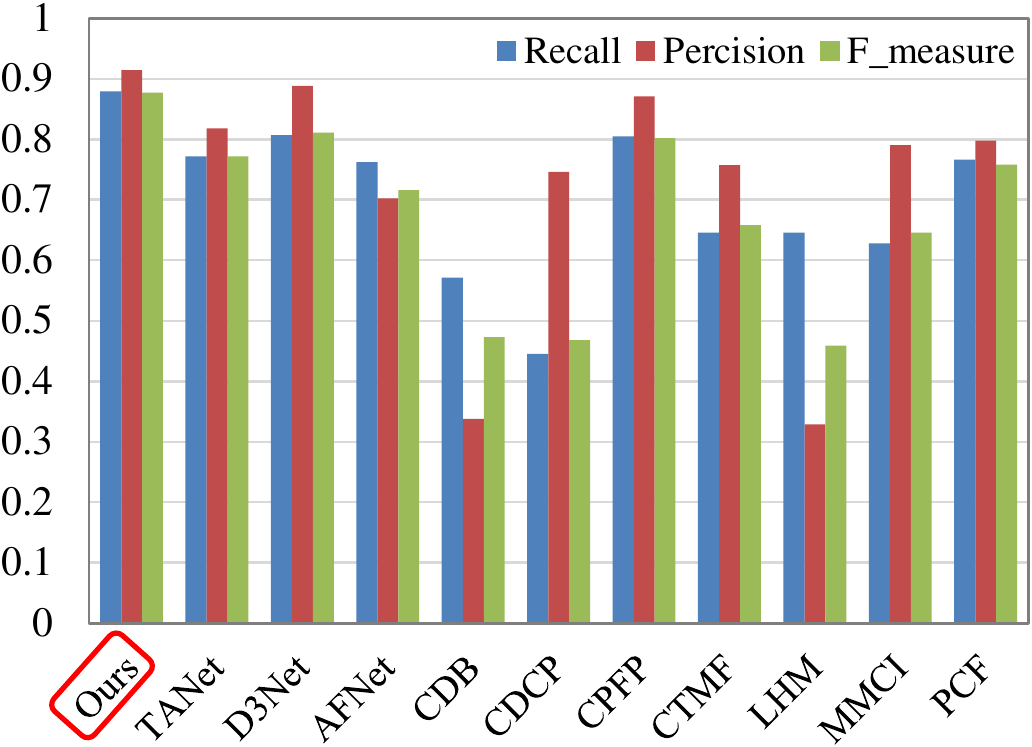}}
		\subfigure[STEREO]{\includegraphics[width=0.235\linewidth]{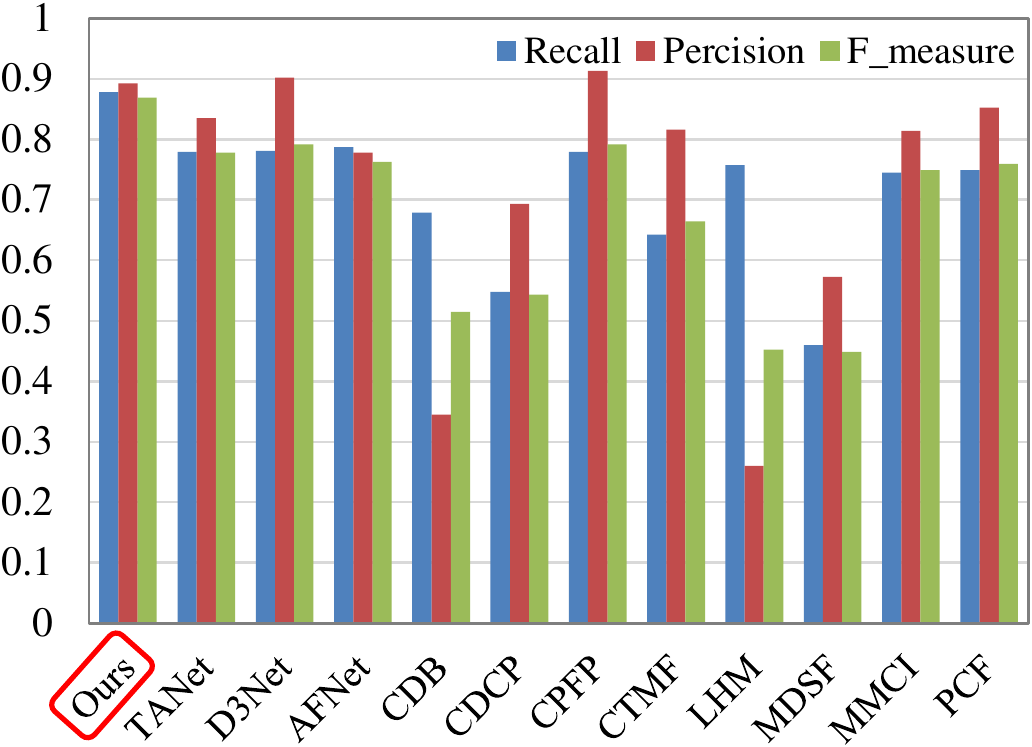}}
	\end{center}
	\caption{Quantitative comparisons. The PR curves (Top) and weighted F-measures (Bottom) of the proposed method and state-of-the-art approaches on four datasets.}
	\label{fig:sub}
\end{figure}
\subsection{Implementation Details}	
Following~\cite{fan2019rethinking,Chen_2018_CVPR,chen2019multi,han2017cnns}, we utilize two VGG-16 networks as the backbones, where one is used for extracting the 2D appearance (RGB) features and the other for extracting 3D geometric (depth) features. We employ the dilated convolutions to ensure that the last two groups of backbones have the same resolution. In the {\em Graph-based Reasoning} (GR) module $g_{\chi}$, three nodes are used in each modality for capturing information of multiple scales, resulting in a graph $\mathcal G$ with six nodes in total. $\mathcal G$ links all nodes of the same modality. For the nodes of different modalities, the edge only connects those nodes with the same scale. During the construction of the {\em Cascade Graph Reasoning} (CGR) module, the features from outputs of the second, third and fifth group of each backbone (different resolutions) are used as inputs for performing cascade graph reasoning. Similar to existing approaches~\cite{Chen_2018_CVPR,chen2019multi,han2017cnns}, BCE loss is used to train our model.

We implement our {\scshape{Cas-Gnn}} using the Pytorch toolbox. The fully equipped model is trained on a PC with GTX 1080Ti GPU for 40 epochs with the mini-batch size of $8$. The input RGB images and depth images are all resized to $256 \times 256$. To avoid overfitting, we perform the following data augmentation techniques: random horizontal flip, random rotate and random brightness. We adopt the Adam with a weight decay of $0.0001$ to optimize the network parameters. The initial learning rate is set to $0.0001$ and the `poly' policy with the power of $0.9$ is used as a mean of adjustment.

\subsection{Ablation Analysis}
In this section, we perform a series of ablations to evaluate each component in our proposed network.

\noindent{\bfseries \small Conventional Feature Fusion vs. Graph-based Reasoning.} To show the effectiveness of graph-based reasoning, we implement a simple baseline model that directly fuses features from the same multi-modality backbones by first performing the concatenate operation and then learning to fuse the learned features for RGB-D based SOD by two $1\times1$ convolutions. Clearly, our graph-based reasoning approach (GR module) achieves much more reliable and accurate results. 

In addition, we further provide two strong baselines to show the superiority of our proposed graph-based reasoning approach. The first one is designed by using the one-shot induced learner (IL)~\cite{bertinetto2016learning,nie2018human} to adapt the learned 3D geometric features to 2D appearance space, making the cross-modality features can be better fused for RGB-D based SOD. The second one uses non-local (NL) module~\cite{wang2018non} to enable 2D appearance feature map to selectively incorporate useful information from 3D geometric features for building powerful representations. As shown in Tab.~\ref{tab:1}, our GR module significantly outperforms these strong baselines. This is because our GR module is capable of explicitly distilling complementary information from 2D appearance (color) and 3D geometry (depth) features while the existing feature fusion approaches fail to reason out high-level relations between them. 
\begin{table}[pt]
	\begin{minipage}{0.37\textwidth}
		\centering
		\caption{Ablation analysis for different graph-related settings.}\label{tab2}
		\resizebox{1\linewidth}{!}{\scriptsize
			\begin{tabular}{l|c|c|c|c|c|c}
				\hline
				\multirow{2}{*}{Methods} &\multicolumn{2}{c|}{\tiny Settings} &\multicolumn{2}{c|}{\tiny NJUD} &\multicolumn{2}{c}{\tiny RGBD135}\\
				\cline{2-7}
				&\tiny $N$ &\tiny $T$ &\tiny $F_\beta$ &\tiny MAE &\tiny $F_\beta$ &\tiny MAE\\
				\hline
				
				\ {\scriptsize {\scshape{Cas-Gnn}}} &2 &3 &0.887 &0.039 &0.890 &0.033\\
				\ {\scriptsize {\scshape{Cas-Gnn}}} &6 &3 &0.903 &0.035 &0.906 &0.028\\
				\ {\scriptsize {\scshape{Cas-Gnn}}} &10 &3 &0.905 &0.035 &0.909 &0.028\\
				\hline
				\ {\scriptsize {\scshape{Cas-Gnn}}} &6 &1 &0.881 &0.038 &0.885 &0.031\\
				\ {\scriptsize{\scshape{Cas-Gnn}}} &6 &3 &0.903 &0.035 &0.906 &0.028\\
				\ {\scriptsize {\scshape{Cas-Gnn}}} &6 &5 &0.907 &0.034 &0.908 &0.028\\
				\hline
				
			\end{tabular}
		}
		\label{tab:2}
	\end{minipage}
	\quad
	\begin{minipage}{0.62\textwidth}
		\centering
		\caption{Ablation analysis on three widely-used datasets.}\label{tab1}
		\resizebox{1\linewidth}{!}{\scriptsize
			\begin{tabular}{l|c|c|c|c|c|c|c|c}
				\hline
				\multirow{2}{*}{\quad Methods} & \multirow{2}{*}{Param.} & \multirow{2}{*}{FLOPs} &\multicolumn{2}{c|}{NJUD~\cite{ju2014depth}} &\multicolumn{2}{c|}{STEREO~\cite{niu2012leveraging}} &\multicolumn{2}{c}{RGBD135~\cite{cheng2014depth}}\\
				\cline{4-9}
				&&&$F_\beta$ &MAE &$F_\beta$ &MAE &$F_\beta$ &MAE\\
				\hline
				
				\quad{\small Baseline} &40.66M&65.64G&0.801 &0.073 &0.813 &0.071 &0.759 &0.052\\
				\quad{\small Baseline + IL} &40.91M&66.21G&0.838 &0.065 &0.841 &0.064 &0.788 &0.046\\
				\quad{\small Baseline + NL} &40.98M&66.86G&0.851 &0.059 &0.852 &0.060 &0.807 &0.043\\
				\quad{\small Baseline + GR (ours)} &41.27M&68.91G& \bf 0.874 & \bf 0.051 & \bf 0.864 & \bf 0.048 & \bf 0.854 & \bf 0.031\\
				\hline
				
				\specialrule{0em}{1pt}{1pt} 
				\quad{\small Baseline + CMFS} &41.88M&72.63G&0.820 &0.068 &0.822 &0.067 &0.780 &0.047\\
				\quad{\small Baseline + HR (ours)} &42.03M&73.19G&0.886 &0.041 &0.871 &0.045 &0.890 &0.033\\
				\quad{\small Baseline + CGR (ours)} &42.28M&73.62G& \bf 0.903 & \bf 0.035 & \bf 0.901 & \bf 0.039 & \bf 0.906 & \bf 0.028\\
				\hline
				
			\end{tabular}
		}
		\label{tab:1}
	\end{minipage}
\end{table}

\noindent{\bfseries \small The Effectiveness of Cascade Graph Reasoning.} A key design of our {\scshape{Cas-Gnn}} is the novel  {\em Cascade Graph Reasoning} module (CGR). To verify the effectiveness of CGR, we use the a common multi-level fusion strategy described in~\cite{Piao_2019_ICCV} (CMFS) for comparison. As shown in Tab.~\ref{tab:1}, our CGR consistently

outperforms CMFS across all datasets. Moreover, our CGR is also superior to the hierarchical reasoning (HR) approach without the {\tt guidance nodes} which is described in Sec.3.3. This indicates that CGR (with the cascade techniques) can better distill and leverage multi-level information than existing strategies. 
\begin{table}[pt]
	\centering
	\caption{\small
		Quantitative comparisons with state-of-the-art methods by S-measure ($S_{\alpha}$), F-measure ($F_{\beta}$), E-measure ($E_{\xi}$) and MAE ($M$) on $7$ widely-used RGB-D
		datasets.
	}
	\resizebox{0.88\textwidth}{!}{
		\begin{tabular}{lr|ccccc|ccccccccc|c}
			\hline\toprule
			&  &\multicolumn{5}{c|}{2014-2017}&\multicolumn{9}{c|}{2018-2020}&\multicolumn{1}{c}{} \\
			\hline
			& Metric 
			&LHM & CDB  & CDCP & MDSF & CTMF & AFNet & MMCI & PCF & TANet& CPFP & D$^3$Net &DMRA &UCNet &ASIF& {\bf Ours} \\
			& & \cite{peng2014rgbd} & \cite{liang2018stereoscopic} & \cite{zhu2017innovative} & \cite{song2017depth}  & \cite{han2017cnns}& \cite{wang2019adaptive} & \cite{chen2019multi} & \cite{Chen_2018_CVPR} &\cite{chen2019three} & \cite{Zhao_2019_CVPR} & \cite{fan2019rethinking} &\cite{Piao_2019_ICCV} &\cite{Zhang2020UCNet} &\cite{li2020asif}&\\
			\midrule
			\midrule
			\multirow{4}{*}{\begin{sideways}\textit{NJUD}\end{sideways}}
			& $S_{\alpha}\uparrow$ &0.514& 0.624 & 0.669 & 0.748 & 0.849 & 0.772  & 0.858 & 0.877 & 0.878 & 0.879 & 0.895 &0.886 &0.897 &0.888& {\bf 0.911}     \\
			& $F_{\beta}\uparrow$  &0.632& 0.648 & 0.621 & 0.775 & 0.845 & 0.775  & 0.852  & 0.872  & 0.874 & 0.877 & 0.889 &0.872 &0.889 &0.900& {\bf 0.903}     \\
			& $E_{\xi}\uparrow$ &0.724& 0.742 & 0.741 & 0.838 & 0.913 & 0.853 & 0.915 & 0.924 & 0.925 & 0.926 & 0.932 &0.908 &0.903&-& {\bf 0.933}    \\
			& $M\downarrow$ &0.205& 0.203 & 0.180 & 0.157  & 0.085 & 0.100 & 0.079  & 0.059  & 0.060 & 0.053 & 0.051 &0.051 &0.043&0.047& {\bf 0.035}     \\
			\midrule
			
			\multirow{4}{*}{\begin{sideways}\textit{STEREO}\end{sideways}}
			& $S_{\alpha}\uparrow$ &0.562& 0.615 & 0.713 & 0.728 & 0.848 & 0.825 & 0.873  & 0.875   & 0.871 & 0.879 & 0.891 &0.886 &{\bf 0.903} &0.868& 0.899 \\
			& $F_{\beta}\uparrow$ &0.683& 0.717 & 0.664 & 0.719 & 0.831 & 0.823 & 0.863 & 0.860 & 0.861 & 0.874 & 0.881 &0.868 &0.885&0.893&{\bf 0.901}  \\
			& $E_{\xi}\uparrow$ &0.771& 0.823 & 0.786 & 0.809 & 0.912 & 0.887 & 0.927& 0.925  & 0.923 & 0.925 & {\bf 0.930} &0.920 &0.922&-& {\bf 0.930} \\
			& $M\downarrow$ &0.172& 0.166 & 0.149 & 0.176 & 0.086 & 0.075 & 0.068  & 0.064 & 0.060 & 0.051 & 0.054 &0.047 &0.040&0.049& {\bf 0.039} \\
			\midrule
			
			\multirow{4}{*}{\begin{sideways}\textit{RGBD135}\end{sideways}}
			& $S_{\alpha}\uparrow$ &0.578& 0.645 & 0.709 & 0.741 & 0.863 & 0.770 & 0.848 & 0.842  & 0.858 & 0.872 & 0.904 &0.901 &-&-& {\bf 0.905} \\
			& $F_{\beta}\uparrow$  &0.511& 0.723  & 0.631 & 0.746 & 0.844 & 0.728 & 0.822  & 0.804   & 0.827 & 0.846 & 0.885 &0.857 &-&-& {\bf 0.906} \\
			& $E_{\xi}\uparrow$ &0.653& 0.830 & 0.811 & 0.851 & 0.932 & 0.881 & 0.928  & 0.893   & 0.910 & 0.923 & 0.946 &0.945 &-&-& {\bf 0.947} \\
			& $M\downarrow$ &0.114& 0.100 & 0.115 & 0.122 & 0.055 & 0.068 & 0.065 & 0.049  & 0.046 & 0.038 & 0.030 &0.029 &-&-& {\bf 0.028} \\
			\midrule
			
			\multirow{4}{*}{\begin{sideways}\textit{NLPR}\end{sideways}}
			& $S_{\alpha}\uparrow$ &0.630& 0.629 & 0.727 & 0.805 & 0.860 & 0.799 & 0.856  & 0.874  & 0.886 & 0.888 & 0.906 &0.899 &0.918&0.884& {\bf 0.919} \\
			& $F_{\beta}\uparrow$  &0.622& 0.618 & 0.645 & 0.793 & 0.825 & 0.771 & 0.815  & 0.841  & 0.863 & 0.867 & 0.885 &0.855 &0.890&0.900& {\bf 0.904} \\
			& $E_{\xi}\uparrow$ &0.766& 0.791 & 0.820 & 0.885 & 0.929 & 0.879 & 0.913  & 0.925  & 0.941 & 0.932 & 0.946 &0.942 &0.951&-& {\bf 0.952} \\
			& $M\downarrow$ &0.108& 0.114 & 0.112 & 0.095 & 0.056 & 0.058 & 0.059  & 0.044  & 0.041 & 0.036 & 0.034 &0.031 &{\bf 0.025}&0.030& {\bf 0.025} \\
			\midrule
			
			\multirow{4}{*}{\begin{sideways}\textit{SSD}\end{sideways}}
			& $S_{\alpha}\uparrow$ &0.566& 0.562 & 0.603 & 0.673 & 0.776 & 0.714 & 0.813 & 0.841 & 0.839 & 0.807 & 0.866 &0.857 &-&-& {\bf 0.872} \\
			& $F_{\beta}\uparrow$ &0.568& 0.592 & 0.535 & 0.703 & 0.729 & 0.687 & 0.781 & 0.807 & 0.810 & 0.766 & 0.847 &0.821 &-&-& {\bf 0.862} \\
			& $E_{\xi}\uparrow$  &0.717& 0.698 & 0.700 & 0.779 & 0.865 & 0.807 & 0.882  & 0.894   & 0.897 & 0.852 & 0.910 &0.892 &-&-& {\bf 0.915} \\
			& $M\downarrow$ &0.195& 0.196 & 0.214 & 0.192 & 0.099 & 0.118 & 0.082  & 0.062   & 0.063 & 0.082 & 0.058 &0.058 &-&-& {\bf 0.047} \\
			\midrule
			
			\multirow{4}{*}{\begin{sideways}\textit{LFSD}\end{sideways}}
			& $S_{\alpha}\uparrow$ &0.553& 0.515 & 0.712 & 0.694 & 0.788 & 0.738 & 0.787  & 0.786 & 0.801 & 0.828 & 0.832 &0.847 &{\bf 0.860}&0.814& 0.849 \\
			& $F_{\beta}\uparrow$ &0.708& 0.677 & 0.702 & 0.779 & 0.787 & 0.744 & 0.771  & 0.775   & 0.796 & 0.826 & 0.819 &0.849 &0.859&0.858& {\bf 0.864} \\
			& $E_{\xi}\uparrow$ &0.763& 0.766 & 0.780 & 0.819 & 0.857 & 0.815 & 0.839  & 0.827   & 0.847 & 0.863 & 0.864 &{\bf0.899} &0.897&-&  0.877 \\
			& $M\downarrow$ &0.218& 0.225 & 0.172 & 0.197 & 0.127 & 0.133 & 0.132  & 0.119   & 0.111 & 0.088 & 0.099 &0.075 &{\bf 0.069}&0.089& 0.073 \\
			\midrule

			\multirow{4}{*}{\begin{sideways}\textit{DUT-RGBD}\end{sideways}}
			& $S_{\alpha}\uparrow$ &0.568& - & 0.687 &- & 0.834 & - & 0.791  & 0.801 & - & - & - &0.888 &-&-& {\bf 0.891} \\
			& $F_{\beta}\uparrow$ &0.659& - & 0.633 &- & 0.792 & - & 0.753  & 0.760   & - & - & - &0.883 &-&-& {\bf 0.912} \\
			& $E_{\xi}\uparrow$ &0.767&- & 0.794 &- & 0.884 & - & 0.855  & 0.858   & - & - & - &0.927 &-&-&  {\bf 0.932} \\
			& $M\downarrow$ &0.174&- & 0.159 & - & 0.097 & - & 0.113  & 0.100 & - & - & - &0.048 &-&-& {\bf0.042} \\
			\midrule

			\hline
		\end{tabular}
		\label{tab:3}
	}
\end{table}

\noindent{\bfseries \small Node Numbers $N$.} To investigate the impact of node numbers $N$ in the GR module, we report the results of our GR module with different $N = 2\cdot n$ in Tab.~\ref{tab:2}. We observe that when more nodes ($n = 1 \mapsto 3$) in each modality are used, the performance of our model improves accordingly. However, when more nodes are included in each modality ($n = 3 \mapsto 5$), the performance improvements are rather limited. This is caused by the redundant information from generated nodes. Therefore, we believe that setting $3$ nodes in each modality ($N=6$) should be a good balance of the speed and accuracy.  

\noindent{\bfseries \small  Message Passing Iterations $T$.} We also evaluate the impact of message passing iterations $T$. As can be seen in Tab.~\ref{tab:2}, when more than three message passing iterations are used for graph reasoning, the model can achieve the best performance. Therefore, we set $T=3$ in our GR module to guarantee a good speed and performance tradeoff.  

\begin{figure}[pt]
	\begin{center}
		\includegraphics[width=0.92\linewidth]{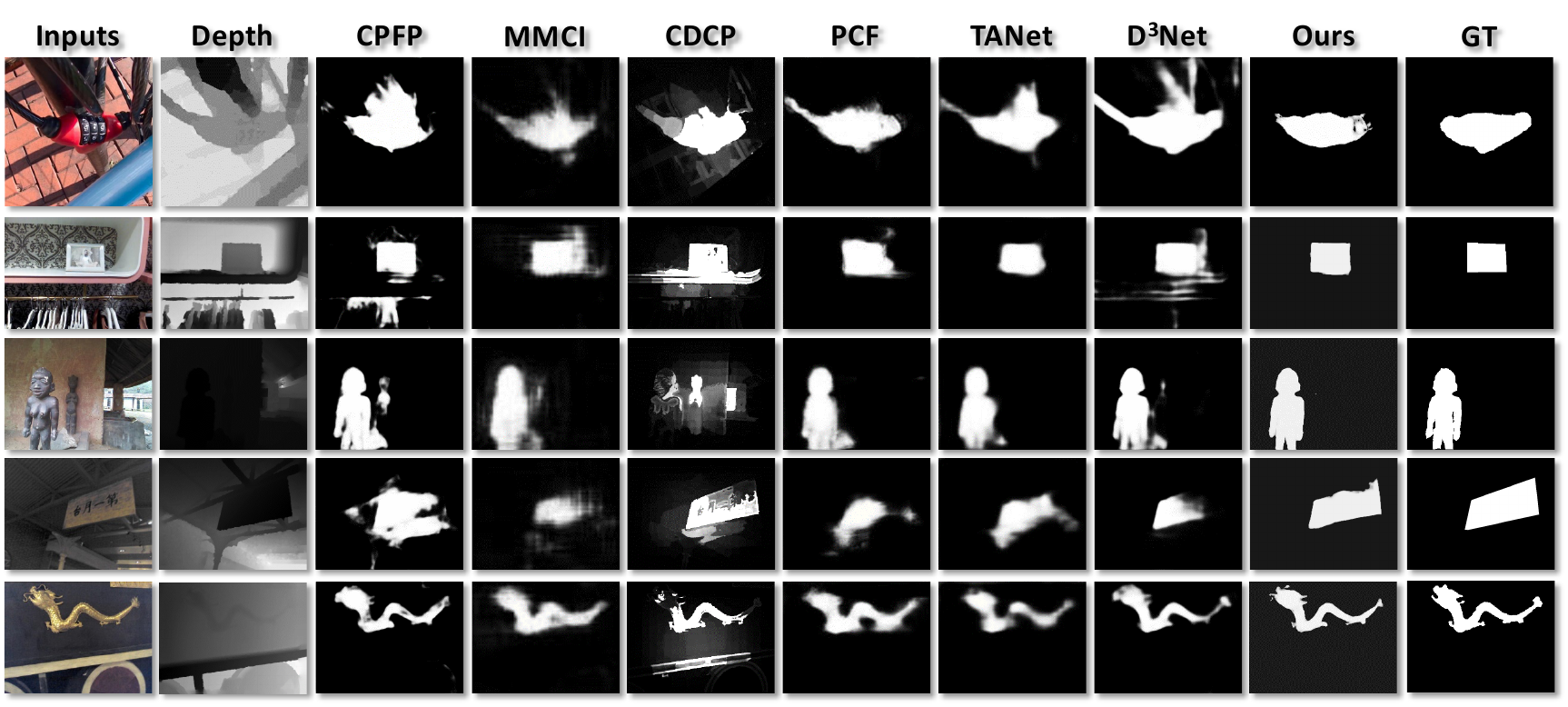}
	\end{center}
	\caption{Qualitative comparisons with state-of-the-art CNNs-based methods.}
	\label{fig:6}
\end{figure}

\subsection{Comparison with SOTAs}

\noindent{\bfseries \small Quantitative Comparisons.} We compare our {\scshape{Cas-Gnn}} with $14$ SOTA models on $7$ widely-used datasets in Tab.~\ref{tab:3}. In general, our {\scshape{Cas-Gnn}} consistently achieves the remarkable performance on all datasets with four evaluation metrics. Clearly, the results demonstrate that explicitly reason and distill mutual beneficial information can help to infer the salient object regions from the clutter images. In addition, we also show the results of widely-used PR curves and weighted F-measure in Fig.~\ref{fig:sub}. As can be seen, our {\scshape{Cas-Gnn}} achieves the best performance on all datasets. All the comparisons with recent SOTAs indicate that mining the high-level relations of multi-modality data sources and perform joint reasoning across multiple feature levels are important, and will largely improve the reliability of deep model for handling cross-modality information. 

\noindent{\bfseries \small Qualitative Comparisons.} Fig.~\ref{fig:6} shows some visual samples of results comparing the proposed {\scshape{Cas-Gnn}}  with state-of-the-art methods. We observe that our {\scshape{Cas-Gnn}} is good at capturing both of the overall salient object regions and local object/region details. This is because our proposed cascade graph reasoning module is able to take both high-level semantics and low-level local details into consideration to build more powerful embeddings for inferring SOD regions.

\section{Conclusion} 
In this paper, we introduce a novel deep model based on graph-based techniques for RGB-D salient object detection. Besides, we further propose to use cascade structure to enhance our GNN model to make it better take advantages of rich, complementary information from multi-level features. According to our experiments, the proposed {\scshape{Cas-Gnn}} successfully distills useful information from both the 2D (color) appearance and 3D geometry (depth) information, and sets new state-of-the-art records on multiple datasets. We believe the novel designs in this paper is important, and can be used to other cross-modality applications, such as RGB-D based object discover or cross-modality medical image analyse. 

\noindent {\bf Acknowledgement}: This research was funded in part by	the National Key R\&D Progrqam of China (2017YFB1302300) and the NSFC (U1613223). 

\clearpage
%
%
\bibliographystyle{splncs04}
\bibliography{egbib}
\end{document}